\documentclass[letterpaper]{article} 
\usepackage{aaai2026}  
\usepackage{times}  
\usepackage{helvet}  
\usepackage{courier}  
\usepackage[hyphens]{url}  
\usepackage{graphicx} 
\usepackage{natbib}  
\usepackage{caption} 
\usepackage{algorithm}
\usepackage{algpseudocode}
\usepackage{xcolor}
\usepackage{booktabs}
\usepackage{multirow}
\usepackage{amsmath}
\usepackage{amssymb}

\usepackage{newfloat}
\usepackage{listings}
\DeclareCaptionStyle{ruled}{labelfont=normalfont,labelsep=colon,strut=off} 
\floatstyle{ruled}
\newfloat{listing}{tb}{lst}{}
\floatname{listing}{Listing}
\title{HIAL: A New Paradigm for Hypergraph Active Learning via Influence Maximization}
\author {
    Yanheng Hou\textsuperscript{\rm 1}\equalcontrib,
    Xunkai Li\textsuperscript{\rm 1}\equalcontrib,
    Zhenjun Li\textsuperscript{\rm 2},
    Bing Zhou\textsuperscript{\rm 3},
    Ronghua Li\textsuperscript{\rm 1},
    Guoren Wag\textsuperscript{\rm 1}
}
\affiliations {
    \textsuperscript{\rm 1}Beijing Institute of Technology, Beijing, China\\
    \textsuperscript{\rm 2}Shenzhen Institute of Technology, Shenzhen, China\\
    \textsuperscript{\rm 3}Shenzhen City Polytechnic, Shenzhen, China\\
    \{houyanheng, lironghuabit\}@126.com, \{cs.xunkai.li, wanggrbit\}@gmail.com 
}

\usepackage{bibentry}
\newtheorem{definition}{Definition}
\newtheorem{theorem}{Theorem}
\newcommand{\cell}[2]{%
  \begin{tabular}{@{}c@{}}#1 \\ \textcolor{red}{#2}\end{tabular}%
}
\begin{document}

\maketitle

\begin{abstract}
In recent years, Hypergraph Neural Networks (HNNs) have demonstrated immense potential in handling complex systems with high-order interactions. However, acquiring large-scale, high-quality labeled data for these models is costly, making Active Learning (AL) a critical technique. Existing Graph Active Learning (GAL) methods, when applied to hypergraphs, often rely on techniques like "clique expansion," which destroys the high-order structural information crucial to a hypergraph's success, thereby leading to suboptimal performance.

To address this challenge, we introduce HIAL (Hypergraph Active Learning), a native active learning framework designed specifically for hypergraphs. We innovatively reformulate the Hypergraph Active Learning (HAL) problem as an Influence Maximization task. The core of HIAL is a dual-perspective influence function that, based on our novel "High-Order Interaction-Aware (HOI-Aware)" propagation mechanism, synergistically evaluates a node's feature-space coverage (via Magnitude of Influence, MoI) and its topological influence (via Expected Diffusion Value, EDV). We prove that this objective function is monotone and submodular, thus enabling the use of an efficient greedy algorithm with a formal (1-1/e) approximation guarantee.

Extensive experiments on seven public datasets demonstrate that HIAL significantly outperforms state-of-the-art baselines in terms of performance, efficiency, generality, and robustness, establishing an efficient and powerful new paradigm for active learning on hypergraphs.
\end{abstract}

%

\section{Introduction}
In recent years, Graph Neural Networks (GNNs)~\cite{kipf2017} have achieved tremendous success in modeling graph-structured data such as social networks, citation networks, and even molecular structures. However, conventional graphs are inherently limited to representing pairwise relationships, failing to capture the higher-order interactions intrinsic to many complex systems. Real-world phenomena, such as multiple authors co-authoring a paper, users participating in group discussions, or various proteins forming a functional complex, are more accurately modeled by hypergraphs~\cite{bianconi2021higher}. This need has significantly propelled the rapid development of Hypergraph Neural Networks (HNNs), which have demonstrated state-of-the-art performance in numerous cutting-edge scientific and industrial applications. Despite the power of HNNs, their development and practical deployment are inseparable from efficient data annotation strategies and principles. The high cost of acquiring large-scale, high-quality labeled data makes Active Learning (AL)~\cite{settles2009} an indispensable and critical technique, aiming to maximize model performance at a minimal annotation cost.

In the domain of graph data, research on Graph Active Learning (GAL) is already quite mature. Early studies primarily focused on strategies based on uncertainty~\cite{settles2009} or data representativeness~\cite{sener2017active}, with subsequent hybrid methods like AGE~\cite{cai2017active} emerging, which combine node attributes with graph structure. However, a key breakthrough in recent years has been the integration of Social Influence Maximization (SIM) theory~\cite{zhanggrain} into the active learning framework. This class of methods has recently garnered widespread attention and is considered one of the most effective approaches. Its remarkable effectiveness stems from an elegant and profound correspondence between the GAL and SIM domains. Specifically, the nodes to be selected for annotation in active learning can be viewed as the \textbf{seed set} in an influence propagation model. Once these seed nodes are labeled, their label information propagates like influence along the graph's topology, \textbf{activating} their neighboring unlabeled nodes. Consequently, the objective of GAL can be redefined as iteratively selecting an optimal set of seed nodes to \textbf{maximize their influence} across the entire graph, thus most efficiently enhancing the overall model performance.

Although GAL has been extensively studied, effectively extending these advanced strategies, especially those based on SIM, to the hypergraph domain remains a significant challenge. The fundamental reason is that existing GAL methods are designed for the pairwise relationships of conventional graphs, and their core algorithms cannot directly comprehend or process the special topological structures of hypergraphs. Consequently, when applying these methods, researchers are forced to rely on techniques like "clique expansion", which "flattens" a hypergraph into a simple graph. However, this process has a fundamental flaw: it destroys the critical higher-order structural information upon which the success of hypergraphs depends~\cite{yadati2019, antelmi2023survey}, inevitably leading to suboptimal query performance. Our empirical analysis further corroborates this finding. Given that hypergraphs, as a higher-order data structure, are demonstrating unique modeling advantages in an increasing number of important fields, designing an efficient and native active learning framework for hypergraph models is both crucial and highly necessary.

To address this challenge, we introduce HIAL, a novel active learning framework designed to operate natively on hypergraphs. We reformulate the Hypergraph Active Learning (HAL) problem as SIM task, aiming to select a seed set of nodes that is both structurally influential and featurally representative. Concretely, HIAL first establishes a High-Order Interaction Aware propagation mechanism to simulate how influence spreads through shared group memberships. Building on this dynamic representation, we construct a dual-perspective influence function that synergizes two complementary metrics: the Magnitude of Influence, which ensures feature-space diversity and coverage, and the Expected Diffusion Value, which quantifies topological impact via probabilistic diffusion. By proving our objective function is monotone and submodular, we employ an efficient greedy algorithm with a formal approximation guarantee of $(1-1/e)$~\cite{nemhauser1978}. 

In summary, our primary contributions are: \((1)\). \textbf{A Novel Paradigm for HAL} We formulate HAL as a influence maximization problem. This shifts the paradigm away from destructive expansion-based methods, preserving the integrity of high-order structures and unlocking superior query performance. \((2)\). \textbf{A Dual-Perspective Influence Framework} We design a sophisticated, theoretically-grounded objective function built upon a novel propagation mechanism. \((3)\). \textbf{High Performance and Generality} We demonstrate the effectiveness, efficiency and generality of HIAL through extensive experiments on seven public datasets.

\section{Preliminary}
\subsection{Notations}
Consider a hypergraph $\mathbf{H} = (\mathcal{V}, \mathcal{E})$, composed of a set of $n = |\mathcal{V}|$ nodes and a set of $m = |\mathcal{E}|$ hyperedges. Each node $v_i \in \mathcal{V}$ is associated with a $d$-dimensional feature vector $\mathbf{x}_i \in \mathbb{R}^d$, and the collection of all node features forms the feature matrix $\mathbf{X} \in \mathbb{R}^{n \times d}$. The ground-truth label for node \(v_i\) is a one-hot vector $\mathbf{y}_i \in \mathbb{R}^{\mathbf{C}}$, where $\mathbf{C}$ is the number of classes. A hyperedge $e_j \in \mathcal{E}$ is a non-empty subset of nodes (i.e., $\emptyset \subset e_j \subseteq \mathcal{V}$), capable of connecting more than two nodes. The structure of the hypergraph is represented by an incidence matrix $\mathbf{H} \in \mathbb{R}^{|\mathcal{V}|\times|\mathcal{E}|}$, where $h_{ve}=1$ if node $v$ belongs to hyperedge $e$, and $h_{ve}=0$ otherwise. The degree of node $v$ is denoted by $d(v) =\Sigma_{e\in \mathcal{E}}h_{ve}$, while the degree of a hyperedge $e$ is denoted by $\delta(e)=\Sigma_{v\in \mathcal{V}}h_{ve}$. We use $\mathbf{D}{v}$ and $\mathbf{D}{e}$ to represent the diagonal matrices of node and hyperedge degrees, respectively.

\subsection{Graph Active Learning}
Active Learning aims to minimize the cost of data labeling by intelligently selecting the most informative unlabeled instances for annotation. Given an unlabeled training set $\mathcal{V}_{\mathrm{train}}$, a labeling budget $\mathcal{B}$, and a machine learning model $\mathcal{M}$ (in our case, an HNN), the goal is to choose a subset $S \subseteq \mathcal{V}_{\mathrm{train}}$ of size $\mathcal{B}$ that, once labeled and used for training, minimizes the model's generalization error on a test set $\mathcal{V}_{\mathrm{test}}$. This can be formally expressed as:
\begin{equation}
\underset{S \subseteq \mathcal{V}_{\mathrm{train}}, |S|=\mathcal{B}}{\operatorname{argmin}} \quad \mathbb{E}_{v_i \in \mathcal{V}_{\mathrm{test}}} \left[ \ell \left( \mathbf{y}_i, P(\hat{\mathbf{y}}_i | \mathbf{x}_i; \mathcal{M}_S) \right) \right]
\end{equation}
where $P(\hat{\mathbf{y}}_i | \mathbf{x}_i; \mathcal{M}_S)$ is the label distribution for node $v_i$ predicted by the model $\mathcal{M}_S$, and $\mathcal{M}_S$ is the HNN trained on the selected labeled set $S$.

While graph active learning is a well-established field, active learning on hypergraphs is significantly less developed. Current HAL research primarily follows two paradigms. Model-dependent methods like ACGNN~\cite{sun2022active} and CGE-AL~\cite{liao2023class} rely on iterative retraining of an HNN to query informative instances. Others, such as $\mathrm{HS}^2$~\cite{chien2019hs}, are designed for clustering tasks and identifying class boundaries sequentially. These approaches are often either tethered to costly, model-dependent iterative loops or designed for specialized query types. In contrast, our work introduces a fundamentally different paradigm. By formalizing data selection as an influence maximization problem, HIAL offers an efficient, non-iterative, and model-independent framework that selects nodes based on their provable influence across both feature and topological domains, addressing a critical gap in active learning for hypergraphs.

\subsection{Social Influence Maximization}
Social Influence Maximization (SIM) problem, first formalized by Kempe et al.~\cite{kempe2003maximizing}, seeks to identify a small subset of "seed" nodes in a network that will maximize the spread of influence. Given a graph $G = (V,E)$ and a budget $\mathcal{B}$, the objective is:
\begin{equation}
\max |\sigma(S)|, \quad \text{s.t.} \quad {S \subseteq V, |S| = \mathcal{B}}
\end{equation}
where $\sigma(S)$ is the set of nodes ultimately activated by the seed set $S$ under a given propagation model (e.g., Linear Threshold or Independent Cascade). While this optimization problem is NP-hard, the influence function $|\sigma(S)|$ is often monotone and submodular. These properties guarantee that greedy algorithm can find a solution with an approximation factor of $(1 - 1/e)$~\cite{nemhauser1978}.

Extending SIM to hypergraphs, however, presents significant theoretical challenges, as direct generalizations of classic diffusion models often result in non-submodular influence functions~\cite{zhu2018social}. This has led to specialized algorithms~\cite{zhu2018social, antelmi2021social} and scalable heuristics~\cite{xie2022influence, auletta2024heuristics}. However, a critical limitation of these frameworks is their exclusive focus on topological data for social network analysis. This leaves a crucial gap for an influence maximization framework tailored to machine learning contexts, where the objective is to select a maximally informative core-set by jointly considering both complex topological structures and rich node features.

\subsection{Hypergraph Neural Networks}
As an extension of GNNs to higher-order data, HNNs learn node representation using hypergraph structures ~\cite{kim2024survey,antelmi2023survey}. The core principle is a two-stage message-passing process: information is first aggregated from nodes to hyperedges, and then propagated back from hyperedges to the nodes. A typical HGNN convolutional layer~\cite{feng2019} can be expressed as:
\begin{equation}
\mathbf{X}^{(k+1)} = \mathbf{D}_v^{-1} \mathbf{H} \mathbf{W} \mathbf{D}_e^{-1} \mathbf{H}^\top \mathbf{X}^{(k)} \mathbf{\Theta}
\end{equation}
where $\mathbf{\Theta}$ is a layer-specific trainable matrix and $\mathbf{W}$ is a diagonal matrix of hyperedge weights. By stacking such layers, HNNs can effectively capture complex higher-order dependencies inherent in the data, enabling downstream tasks such as node classification and link prediction.

\section{The HIAL Framework}
\begin{figure}
\centering
    \includegraphics[width=1\linewidth]{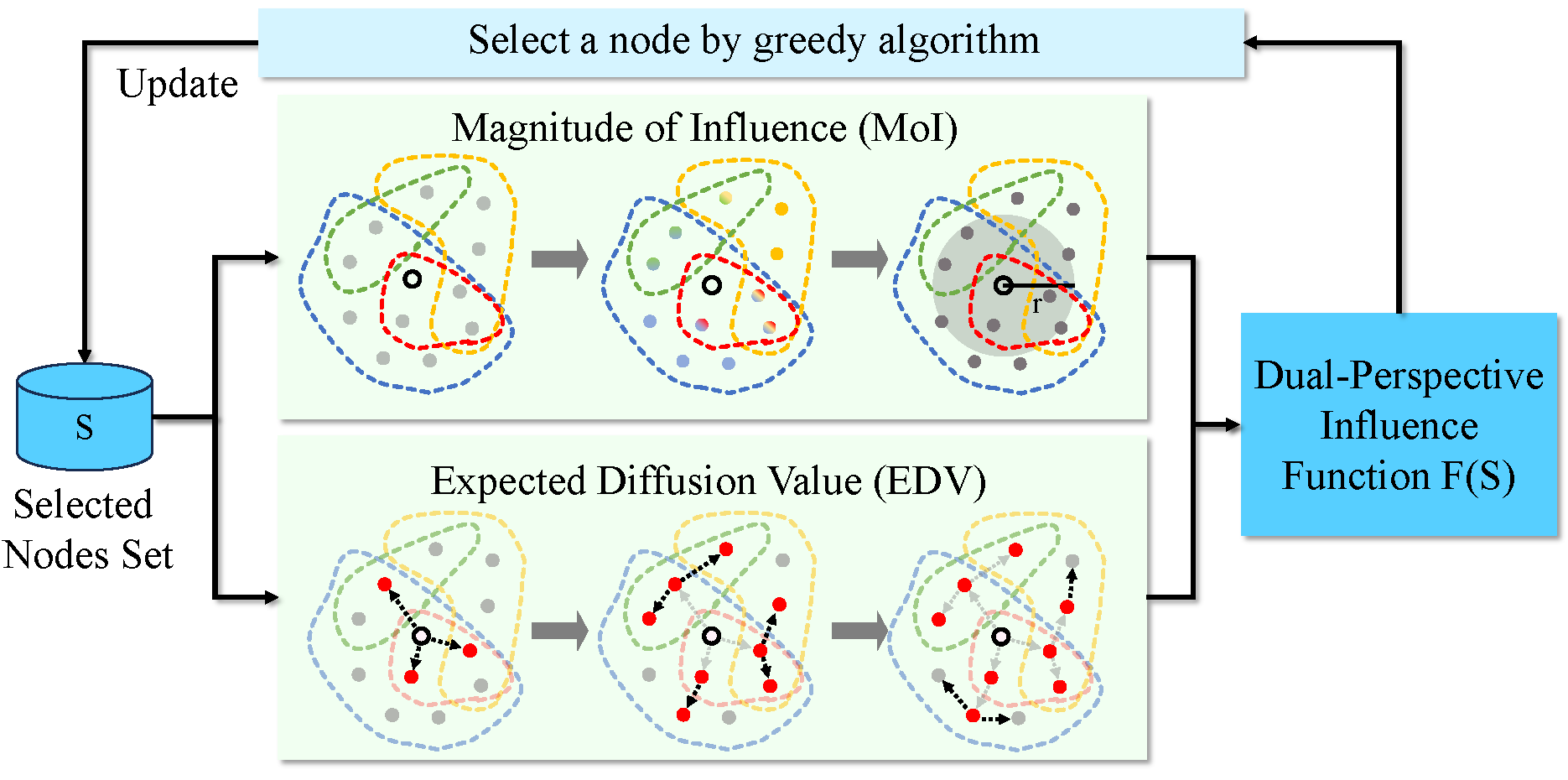}
\caption{The workflow of HIAL}
\label{workflow}
\end{figure}

In this section, we present the details of the HIAL framework, illustrated in Figure~\ref{workflow}. Given a hypergraph \(\mathbf{H} = (\mathcal{V}, \mathcal{E})\) and a node feature matrix \(\mathbf{X}\), our method proceeds in two main stages. First, we establish a novel, parameter-free feature propagation mechanism that is aware of high-order interactions. Second, based on this propagation, we use a dual-perspective influence function as maximization objective, which is then optimized using a greedy algorithm to select the most informative nodes.

\subsection{Influence Propagation Model}
The effectiveness of Hypergraph Neural Networks primarily stems from their ability to aggregate information across hypergraph structure. Recent studies suggest that this power is mainly derived from feature smoothing over hypergraph neighborhoods rather than complex non-linear transformations~\cite{rossi2020sign,he2020lightgcn}. Generalizing this insight, we first establish a parameter-free propagation mechanism to serve as the foundation for quantifying influence.\\
\textbf{\underline{Decoupled HNN Propagation}}
A standard HNN propagation layer, once decoupled from learnable parameters, often takes the form $\mathbf{X}^{(k+1)} = \mathbf{T} \mathbf{X}^{(k)}$, where $\mathbf{T}$ is a transition matrix derived from the hypergraph structure (e.g., $\mathbf{T} = \mathbf{D}_v^{-1/2} \mathbf{H} \mathbf{W} \mathbf{D}_e^{-1} \mathbf{H}^T \mathbf{D}_v^{-1/2}$). However, such formulations typically treat all connections uniformly. They either fail to distinguish between interactions occurring within different hyperedges or implicitly reduce high-order structures to a series of pairwise links (a "clique expansion"), which obfuscates the underlying group interaction dynamics.\\
\textbf{\underline{HOI-Aware Propagation}}
To overcome these limitations, we design a propagation mechanism that explicitly respects the nature of high-order interactions. Our approach is inspired by recent advancements in modeling random walks on hypergraphs~\cite{carletti2020random}, which posit that a hyperedge represents a fundamental unit of interaction. In such models, information is more likely to flow between nodes that belong to the same hyperedge, reflecting that real-world interactions are inherently group-based.

To conduct this, we construct a \textbf{HOI-Aware} transition matrix. Instead of uniform weights, we assign a higher weight to pairs of nodes that frequently co-occur across many shared hyperedges. Specifically, we quantify the interaction strength between node $i$ and node $j$ by counting their "common neighbors" within the shared hyperedges.

\begin{definition}
The "common neighbor" count of node $j$ for node $i$ is defined as:
\begin{equation}\label{eq:cnc}
l_{ij} = 
\sum_{\alpha\in E} h_{i\alpha} h_{j\alpha} (C_{\alpha\alpha} - 1),\quad i\neq j 
\end{equation}
where $h_{i\alpha}$ is the entry in the incidence matrix and $C_{\alpha\alpha} = \sum_{k\in V} h_{k\alpha}^2$ is the degree of hyperedge $\alpha$.
\end{definition}

The term $h_{i\alpha} h_{j\alpha}$ ensures we only consider hyperedges $\alpha$ containing both $i$ and $j$. The term $(C_{\alpha\alpha} - 1)$ then counts the number of other nodes within that shared context. Thus, $l_{ij}$ is a multiplicity-aware measure that captures the strength of the relationship between $i$ and $j$ by summing up the sizes of their shared interaction groups. We then normalize the resulting matrix $L = (l_{ij})$ to form a symmetric transition matrix $\tilde{L}$ and combine it with the identity matrix to preserve self-information:
\begin{definition}
The high-order interaction aware propagation of node feature is defined as:
\begin{equation}\label{eq:HOIprop}
\mathbf{X}^{(k+1)} = \alpha \tilde{L}\mathbf{X}^{(k)} + (1 - \alpha)\mathbf{X}^{(0)}
\end{equation}    
\end{definition}
where $\alpha$ is a trade-off hyperparameter. This construction ensures that feature propagation preferentially flows along paths of strong, high-order interaction.

\subsection{Influence Function}
\textbf{\underline{Magnitude of Influence}}
Under the given decoupled propagation mechanism, \(\mathbf{X}^{(k)}\) actually captures the information from its \(k\)-hop neighbors. By taking feature propagation as a type of influence propagation, we open up a new perspective for the HAL problem from the viewpoint of influence maximization. Inspired by ~\cite{wang2020unifying},~\cite{xu2018representation} and ~\cite{zhanggrain}, we measure the influence of node \(i\) on node \(j\) after \(k\)-step propagation by the impact of \(i\)'s initial feature on \(j\)'s \(k\)-step propagated feature:
\begin{definition}
The feature influence of node \(i\) on node \(j\) after \(k\)-step propagation is defined as the \(L_1\) norm of the expectation of the Jacobian matrix:
\begin{equation}\label{influ}
I(j, i, k) = \left\Vert E\left[ {\partial X_j^{(k)}}/{\partial X_i^{(0)}} \right] \right\Vert_1
\end{equation}
The normalized feature influence is:
\begin{equation}\label{ninflu}
I_j(i, k) = \frac{I(j, i, k)}{\sum_{l\in V} I(j, l, k)}
\end{equation}
\end{definition}
For a node set \(S\), the set-to-point influence on node \(j\) is the maximum influence from any node in \(S\): \(I_j(S, k) = \max_{u \in S} I_j(u, k)\) A node \(j\) is considered "activated" by \(S\) if this influence exceeds a threshold \(\theta\). Formally:
\begin{definition}
Given a threshold \(\theta\), \(k\)-step feature propagation, and a seed set \(S\), the activated node set \(\sigma(S)\) is:
\begin{equation}\label{eq:activeset}
\sigma(S) = \{j \in V \mid I_j(S, k) > \theta\}
\end{equation}
\end{definition}
The function $|\sigma(S)|$ is provably non-decreasing and submodular as followed:
\begin{theorem}\label{theo1}
$|\sigma(S)|$ is non-decreasing and submodular function, i.e. $\forall S \subseteq T$, $|\sigma(T)|\geq |\sigma(S)|    
$ and $|\sigma(S)\cup\{v\}| - |\sigma(S)| \geq |\sigma(T)\cup\{v\}|-|\sigma(T)| $
\end{theorem}

\begin{figure}
\centering
    \includegraphics[width=1\linewidth]{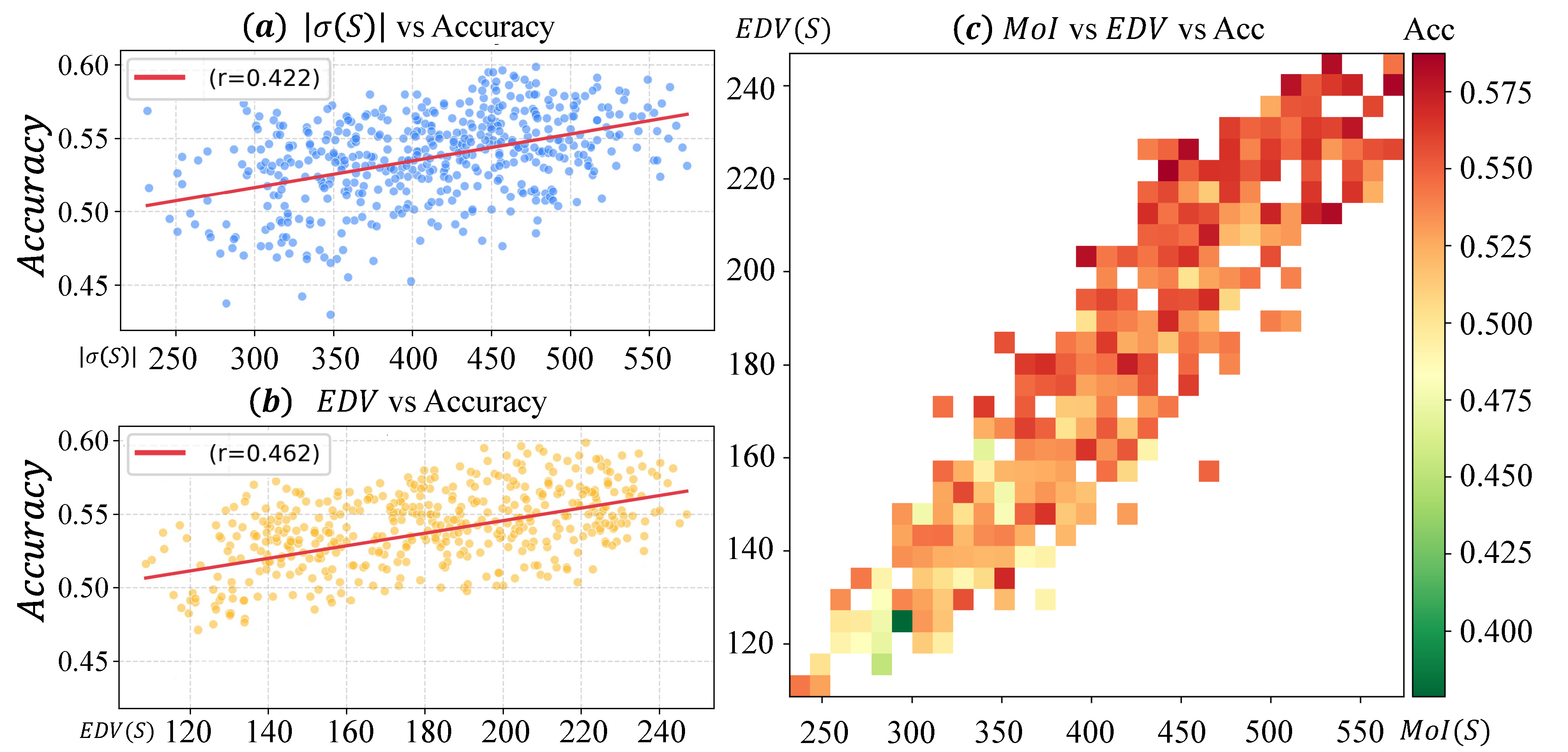}
\caption{The relationship between seed set \(S\) (\(|S|=20\mathbf{C}\)) and test accuracy of HNN model trained on \(S\) of Cora.} 
\label{heatmap}
\end{figure}

As previously discussed, we transform HAL into solving the SIM problem on hypergraphs. An intuitive idea is that nodes capable of influencing more unlabeled nodes are more valuable and influential in the hypergraph. To intuitively verify this hypothesis, we randomly selected seed node sets of same size $(20\mathbf{C})$ for labeling and directly used them to train the downstream HNN model. We plotted the relationship between the influence magnitude $\sigma(S)$ of these seed node sets and the performance of the HNN model, as shown in Figure \ref{heatmap}. As the scale of nodes activated by labeled set increases, the accuracy of the HNN model also increases, indicating a positive correlation between them.

However, Figure~\ref{heatmap}\((a)\) also reveals a crucial subtlety: seed sets with the same activation scale $|\sigma(S)|$ can lead to significantly different model accuracies. This indicates that simply maximizing the number of activated nodes is insufficient. The reason is that HNN aggregation operations capture information in feature space. It is a reasonable assumption that nodes close to each other in the feature space are likely to share the same label. When a node is "activated" via propagation, its influence should also extend to its feature-space neighbors. Therefore, to better capture the total value of an activated set, we must account for this indirect influence. We model this assuming that the influence of an activated node $u$ covers a ball $G_u$ of radius $r$ in the feature space, where $G_u = \{v \in V \mid d(X_u^{(k)}, X_v^{(k)}) \leq r\}$. This leads to our refined influence function.
\begin{definition}
The magnitude of influence \(MoI\) of a seed set \(S\) is defined as:
\begin{equation}\label{MoI}
MoI(S) = \left| \bigcup_{u\in\sigma(S)} G_u \right|
\end{equation}
\end{definition}
When \(r = 0\), \(MoI(S)\) reduces to \(|\sigma(S)|\). Similar to \(\sigma(S)\), \(MoI(S)\) exhibits desirable properties:
\begin{theorem}\label{theo2}
\(MoI(S)\) is a non-decreasing and submodular function with respect to \(S\).
\end{theorem}

\begin{figure*}[t]
\centering
    \includegraphics[width=1\linewidth]{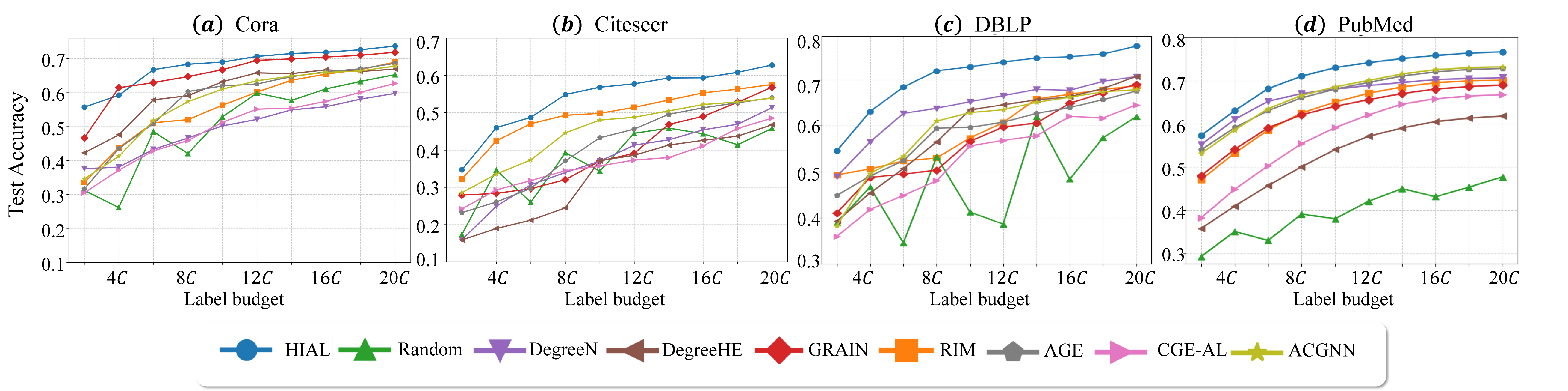}
    \caption{Test accuracy \% across different label budget \(\mathcal{B}\) from \(2\mathbf{C}\) to \(20\mathbf{C}\) on four cocitation datasets} 
\label{acc with budget}    
\end{figure*}

\noindent
\textbf{\underline{Expected Diffusion Value}} While \(MoI\) effectively captures influence in the feature space, it focuses primarily on feature neighborhoods. However, it can overlook crucial information embedded within the intrinsic topological propagation patterns of the hypergraph. The true impact of a node depends not only on its feature, but also on its strategic position and potential to propagate influence through the network structure. To create a truly holistic selection strategy, we must also quantify how the influence spreads through the high-order connections themselves.

Directly simulating influence cascades on hypergraphs, as done in traditional SIM problems, is computationally prohibitive due to the combinatorial complexity of high-order interactions~\cite{wang2024hedv}. To overcome this challenge, we introduce the \textbf{Expected Diffusion Value (\(EDV\))}. The goal of \(EDV\) is to quantify the topological influence of a seed set $S$ by explicitly modeling the probabilistic diffusion through hyperedges, complementing \(MoI\) with topological awareness. The core motivation lies in its ability to characterize the "collective activation" mechanism unique to hypergraphs. Unlike in pairwise graphs, a hyperedge $e$ enables a single seed node $u \in S$ to simultaneously attempt to activate all other members of $e$. The \(EDV\) formula elegantly models this local, one-hop diffusion process.

\begin{definition}
The expected diffusion value \(EDV\) of a seed set $S$ is defined as:
\begin{equation}\label{eq:EDV}
EDV(S) =|S|+\sum_{v \in N(S)} \left[1 - \prod_{u \in S \cap \mathbf{N}(v)} \left(1 - \beta \cdot \frac{n_{u,v}}{n_u}\right)\right]
\end{equation}
where $|S|$ accounts for the initial seeds, $N(S)$ is the set of one-hop neighbors of $S$, $\mathbf{N}(v)$ is the set of neighbors of $v$, $n_{u,v}$ is the number of hyperedges shared by $u$ and $v$, $n_u$ is the hyperdegree of $u$, and $\beta \in (0,1)$ is a base propagation probability.
\end{definition}

The intuition behind Eq.~\eqref{eq:EDV} is as follows: for each non-seed neighbor $v \in N(S)$, the term $(1 - \beta \cdot \frac{n_{u,v}}{n_u})$ represents the probability that a specific seed $u$ \textit{fails} to activate $v$. The product $\prod(\cdot)$ thus computes the probability that $v$ evades activation from \textit{all} its seed neighbors in $S$. Consequently, one minus this product gives the total probability that $v$ is activated by the set $S$. Summing this over all neighbors provides the expected number of newly influenced nodes in the local vicinity of $S$. Like $MoI(S)$, this function exhibits desirable properties for optimization:
\begin{theorem}\label{theo3}
$EDV(S)$ is a non-decreasing and submodular function with respect to $S$.
\end{theorem}
\textbf{\underline{Discussions}} 
The necessity of integrating \(EDV\) is not merely a theoretical construct but is also strongly validated by our empirical results, as shown in Figure \ref{heatmap}(\(b\)) and (\(c\)). For any fixed level of \(MoI\), the accuracy consistently increases with higher \(EDV\). This finding provides clear evidence that \(EDV\) captures a dimension of influence that is both complementary and critical to \(MoI\). Therefore, integration of $EDV(S)$ enriches HIAL by explicitly modeling high-order interactions critical for collective influence. This dual-perspective design ensures that HIAL selects seed nodes that maximize both the diversity of coverage (via \(MoI\)) and the breadth of structural influence (via \(EDV\)), laying a solid foundation for effective active learning on hypergraphs.\\
\textbf{\underline{Unified Objective Function}}
We combine these two perspectives into a single objective function $F(S)$:
\begin{equation}\label{eq:FS}
F(S) = \gamma \cdot \frac{MoI(S)}{\hat{MoI}} + (1-\gamma) \cdot \frac{EDV(S)}{\hat{EDV}}
\end{equation}
where $\gamma \in [0,1]$ is a trade-off parameter and $\hat{MoI}$, $\hat{EDV}$ are normalization factors. 
This linear combination preserves the crucial properties of monotonicity and submodularity. 

\subsection{Selection Algorithm}
\textbf{\underline{Greedy Algorithm}}
Algorithm~\ref{alg:greedy} shows a sketch of our greedy node selection method for HNNs. Without losing generality, we consider a batch setting with \(\mathcal{B}\) rounds where one node is selected in each iteration. Given the layer \(k\) , we first perform the HOI-Aware propagation. Once a node \(v\) is selected and added to \(S\), we update \(MoI(S\cup\{v\})\) and \(EDV(S\cup\{v\})\) according to Eq.~\eqref{MoI} and Eq.~\eqref{eq:EDV}, respectively. Next, we select the node \(v^*\) generating the maximum marginal gain, and the selected nodes set \(S\) are updated. For monotone and submodular \(F(S)\), the final selected node set \(S\) is within a factor of (\(1-1/{e}\)) of the optimal set \(S^*\): \(F(S) \geq (1-1/e)F(S^*)\).

\begin{algorithm}[t]
\caption{Greedy Node Selection for Hypergraphs}
\label{alg:greedy}
\textbf{Input:} \leavevmode\hangindent=3em \hangafter=1 Hypergraph $\mathbf{H} = (\mathcal{V}, \mathcal{E})$, node feature $\mathbf{X}$, propagation mechanism \(f\) and layer number \(k\), labeling budget $\mathcal{B}$.

\textbf{Output:} Seed set $S$

\begin{algorithmic}[1]
\For{$i = 1,2,...,k$}
    \State $X^{(i)} = f(X^{(i-1)}, \mathbf{T}, X^{(0)})$;
\EndFor

\State $S = \emptyset$;
\State $\hat{EDV} = EDV(\mathcal{V})$; 
\State $\hat{MoI} = MoI(\mathcal{V})$;

\For{$t = 1,2,...,\mathcal{B}$}
    \For{$v \in \mathcal{V}_{\text{train}} \setminus S$}
        \State Update $MoI(S \cup \{v\})$ by Eq.\eqref{MoI};
        \State Update $EDV(S \cup \{v\})$ by Eq.\eqref{eq:EDV};
    \EndFor
    \State $v^*= \text{argmax}_{v\in \mathcal{V}_{\text{train}}\setminus S}F(S\cup\{v\})-F(S)$;
    \State $S = S \cup \{v^*\}$; 
\EndFor\\
\Return $S$
\end{algorithmic}
\end{algorithm}

\begin{table*}[t]
\caption{Performance comparison (Mean accuracy \% ± std) of active learning methods on HNNs for node classification. The label budget $\mathcal{B}=20\mathbf{C}$ for small datasets and $50\mathbf{C}$ for Recipe200K. Hightlighted are the \textcolor{red}{top}, \textcolor{blue}{second}, \textcolor{orange}{third} accuracy.}
\centering
\begin{tabular}{c|ccccccccc}
\toprule
Method & Cora & Citeseer & DBLP & PubMed & Yelp3K & Mushroom & Recipe200K \\
\midrule
Random & 65.27 ± 3.46 & 45.77 ± 5.36 & 61.97 ± 1.89 & 47.77 ± 0.29 & 61.62 ± 0.00 & 55.00 ± 1.41 & 56.07 ± 0.19\\
Degree-N & 59.78 ± 1.13 & 51.32 ± 1.38 & 66.68 ± 0.73 & 70.01 ± 1.85 & 69.45 ± 0.15 & 73.25 ± 0.44 & 54.26 ± 0.59\\
Degree-HE & 66.91 ± 0.72 & 46.73 ± 1.24 & 65.71 ± 0.67 & 61.82 ± 0.88 & 64.07 ± 5.22 & 70.87 ± 0.12 & 53.24 ± 0.11\\
\midrule

AGE & 68.64 ± 1.68 & 53.98 ± 1.47 & 67.53 ± 2.34 & \textcolor{orange}{72.82 ± 2.05} & \textcolor{orange}{98.88 ± 0.54} & 63.44 ± 1.18 & OOT\\
CGE-AL & 62.73 ± 1.70 & 48.51 ± 2.73 & 64.51 ± 2.43 & 66.76 ± 2.56 & 98.12 ± 1.17 & 77.86 ± 4.28 & 55.60 ± 0.38\\
ACGNN & 67.81 ± 1.73 & 53.87 ± 1.96 & 67.83 ± 2.01 & \textcolor{blue}{73.20 ± 1.73} & 96.87 ± 0.10 & 90.88 ± 8.49 & \textcolor{orange}{56.93 ± 1.20}\\
\midrule
GRAIN & \textcolor{blue}{71.90 ± 0.96} & \textcolor{orange}{56.81 ± 0.40} & \textcolor{blue}{68.92 ± 1.21} & 68.97 ± 0.59 & \textcolor{blue}{98.90 ± 0.18} & \textcolor{orange}{94.51 ± 3.76} & \textcolor{blue}{57.86 ± 0.98}\\
RIM & \textcolor{orange}{68.99 ± 0.73} & \textcolor{blue}{57.54 ± 0.74} & \textcolor{orange}{68.68 ± 0.82} & 70.13 ± 1.22 & 96.25 ± 0.11 & \textcolor{blue}{95.50 ± 1.88} & OOT\\
\midrule
\textbf{HIAL} & \textcolor{red}{\textbf{73.72 ± 0.41}} & \textcolor{red}{\textbf{62.76 ± 0.35}} & \textcolor{red}{\textbf{77.30 ± 0.37}} & \textcolor{red}{\textbf{76.66 ± 0.72}} & \textcolor{red}{\textbf{99.62 ± 0.07}} & \textcolor{red}{\textbf{96.35 ± 6.31}} & \textcolor{red}{\textbf{59.13 ± 1.09} }\\

\bottomrule
\end{tabular}
\label{Accuracy result}
\end{table*}

\section{Experiments}
To validate the effectiveness and efficiency of the HIAL framework for hypergraph data selection, we design experiments to address the following key questions:\\
\textbf{Q1: Effectiveness.} Does HIAL outperform state-of-the-art baselines in hypergraph active learning (HAL) ?\\
\textbf{Q2: Interpretability.} Do the core components of HIAL (\(MoI\), \(EDV\), and HOI propagation) contribute to its performance?\\
\textbf{Q3: Efficiency and Scalablity.} Is HIAL efficient and scalable for large-scale hypergraphs?\\
\textbf{Q4: Generalization.} Can HIAL perform well with different HNN models?\\
\textbf{Q5: Robustness.} How sensitive is HIAL to hyperparameters? 
\subsection{Experimental Settings} 
\textbf{\underline{Datasets}}
We evaluate our model on four citation networks: Cora, Citeseer, PubMed, DBLP, social network: Yelp3K, biological network: Mushroom and large scale network: Recipe200K. More description is given in Appendix A.\(2\).\\
\textbf{\underline{Baselines}}
We compare HIAL with heuristic methods (Random, Degree-N, Degree-HE), learning-based methods AGE~\cite{hu2021adaptive}, CGE-AL~\cite{liao2023class}, ACGNN~\cite{sun2022active}, and learning-free methods GRAIN~\cite{zhanggrain}, RIM~\cite{zhang2021rim}. More details can be found in Appendix A.\(3\).\\
\textbf{\underline{HNN models}}
We conduct experiments using HIAL equipped with the propagation of existing HNN models and demonstrate the generalization of HIAL in Q\(4\). The description of these GNNs is provided in Appendix A.\(4\).\\
\textbf{\underline{Settings}}
For each method, we use the hyper-parameter tuning toolkit or follow the original papers to find the optimal hyperparameters. We repeat each method ten times and report the mean performance ± std. The parameter settings and implementation detail can be found in Appendix A.\(5\).

\subsection{Performance comparison (Q1)}
\begin{table}[t]
\centering
\caption{Ablation study at label budget \(\mathcal{B}=20\mathbf{C}\). The \textcolor{red}{gap} from original method is highligted.}
\begin{tabular}{c|cccc}
\toprule
Method & Cora & Citeseer & DBLP & PubMed \\
\midrule
\textbf{HIAL} & \textbf{73.72} & \textbf{62.76} & \textbf{77.30} & \textbf{76.66} \\
\midrule
w/o \(MoI\) & \cell{70.23}{-3.49} & \cell{55.78}{-6.98} & \cell{73.61}{-3.69} & \cell{72.45}{-4.21} \\
\midrule
w/o \(EDV\) & \cell{70.79}{-2.93} & \cell{60.22}{-2.54} & \cell{75.66}{-1.64} & \cell{73.80}{-2.86} \\
\midrule
w/o \(HOIP\) & \cell{68.84}{-4.88} & \cell{56.47}{-6.29} & \cell{66.75}{-10.55} & \cell{70.52}{-6.14} \\
\bottomrule
\end{tabular}
\label{Ablation}
\end{table}
We evaluate the performance of HIAL against baselines by analyzing the test accuracy of the trained HNN model across a range of labeling budgets, from $2\mathbf{C}$ to $20\mathbf{C}$. Figure~\ref{acc with budget} clearly illustrates that HIAL not only achieves the highest accuracy at all budget levels but also exhibits a significantly steeper learning curve. This demonstrates that our method identifies more informative nodes early in the process, leading to rapid performance gains. The practical benefit of this is a substantial reduction in labeling cost. For instance, on the Citeseer dataset, HIAL achieves over 55\% with just 48 labeled nodes, a performance level that requires the competitive baseline RIM to label 108 nodes. This represents a \textbf{labeling cost reduction of over 50\%} for reaching the same performance threshold, directly validating the efficacy of our proposed influence function.

As shown in Table~\ref{Accuracy result}, HIAL consistently achieves SOTA across all seven datasets when label budget \(\mathcal{B}=20\mathbf{C}\). By holistically integrating \(MoI\) with \(EDV\), HIAL surpasses all baselines. On the  citation networks, HIAL improves upon the strongest learning-free baselines, outperforming GRAIN by \textbf{1.82\% to 8.38\%} and RIM by \textbf{4.73\% to 8.62\%}. Unlike learning-based methods (e.g., AGE, CGE-AL, ACGNN) whose selection criteria are dependent on a model, HIAL is guided by its stable, dual-perspective influence function. This advantage is prominent in Table~\ref{Accuracy result}: on Citeseer, HIAL outperforms the learning-based ACGNN by a commanding margin of \textbf{8.89\%}. This consistent and significant lead underscores the power of our method in selecting the most valuable nodes for model training.

\begin{figure}
    \centering
    \includegraphics[width=1\linewidth]{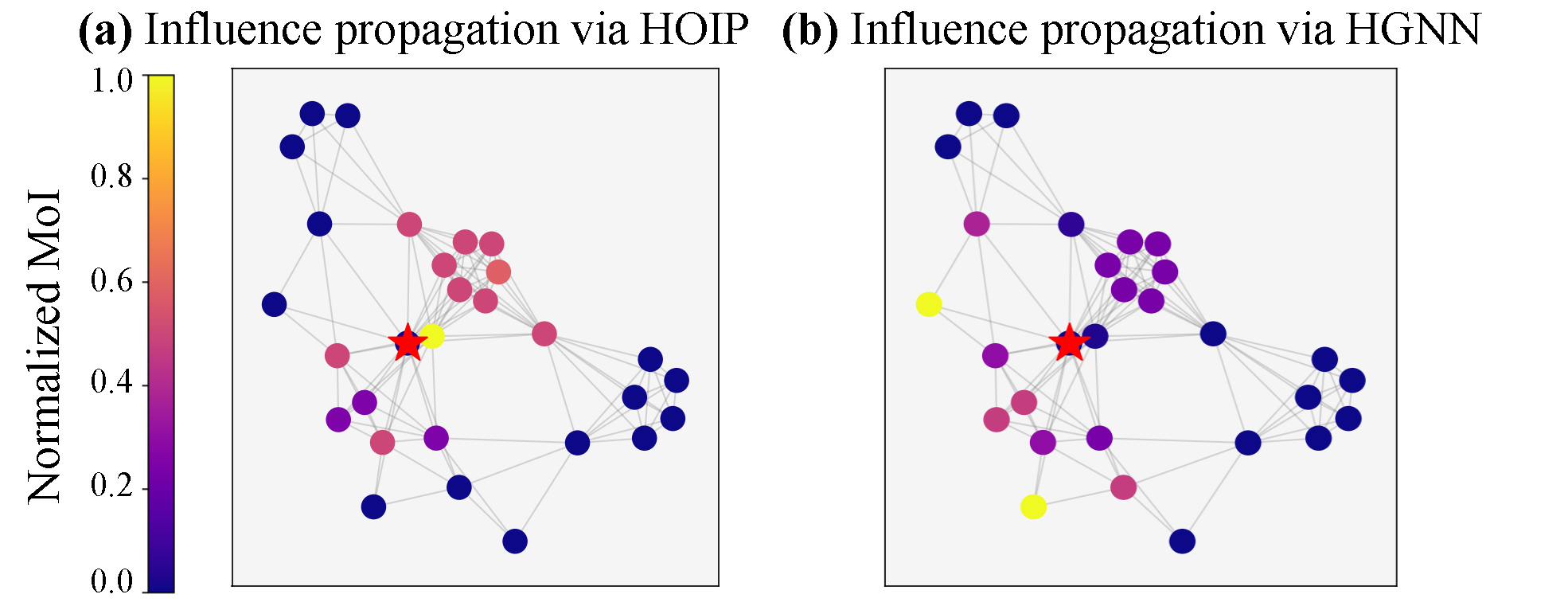}
    \caption{Distribution of Normalized \(MoI(\{v^*\}\cup\{v\})\) via various propagation. \(v^*\) is marked with red star.}
    \label{HOIP}
\end{figure}

\subsection{Model Interpretability(Q2)}
To evaluate each component of the HIAL framework, we conducted a comprehensive ablation study isolating the impact of the High-Order Interaction Aware Propagation (HOIP) module, the Magnitude of Influence (\(MoI\)), and the Expected Diffusion Value (\(EDV\)). The results are summarized in Table \ref{Ablation}.\\
\textbf{\underline{Excluding \(MoI\)}} Relying exclusively on EDV caused a significant performance degradation across all datasets, with accuracy reductions up to 7.1\%. This underscores the necessity of \(MoI\) for assessing feature-space diversity; without it, the model selects topologically central but semantically redundant nodes, impairing performance.\\
\textbf{\underline{Excluding \(EDV\)}} Conversely, relying only on MoI also resulted in a consistent performance decline. This validates the critical role of modeling topological propagation, as neglecting the \(EDV\) prevents the model from capitalizing on crucial high-order pathways for effective influence spread.\\
\textbf{\underline{Disabling Propagation}} Disabling the HOIP module and computing MoI from the static initial feature matrix induced the most substantial performance collapse. This decisively establishes \(HOIP\) as the foundational component of HIAL. Its absence leads to an influence assessment based on less informative representation, fundamentally misjudging the influence potential that arises from network interactions.\\
\textbf{\underline{Model Visualization}}
To illustrate \(HOIP\)'s effectiveness, we conducted a visualization experiment on a sample from the Cora dataset. As shown in Figure ~\ref{HOIP}, \(HOIP\) accurately presents a hierarchical influence distribution that identifies key collaborators. In contrast, a standard HGNN model produces a flat and indistinct result due to information loss, failing to distinguish node importance.

\subsection{Efficiency and Scalability(Q3)}

\begin{figure}[t]
    \centering
    \includegraphics[width=1\linewidth]{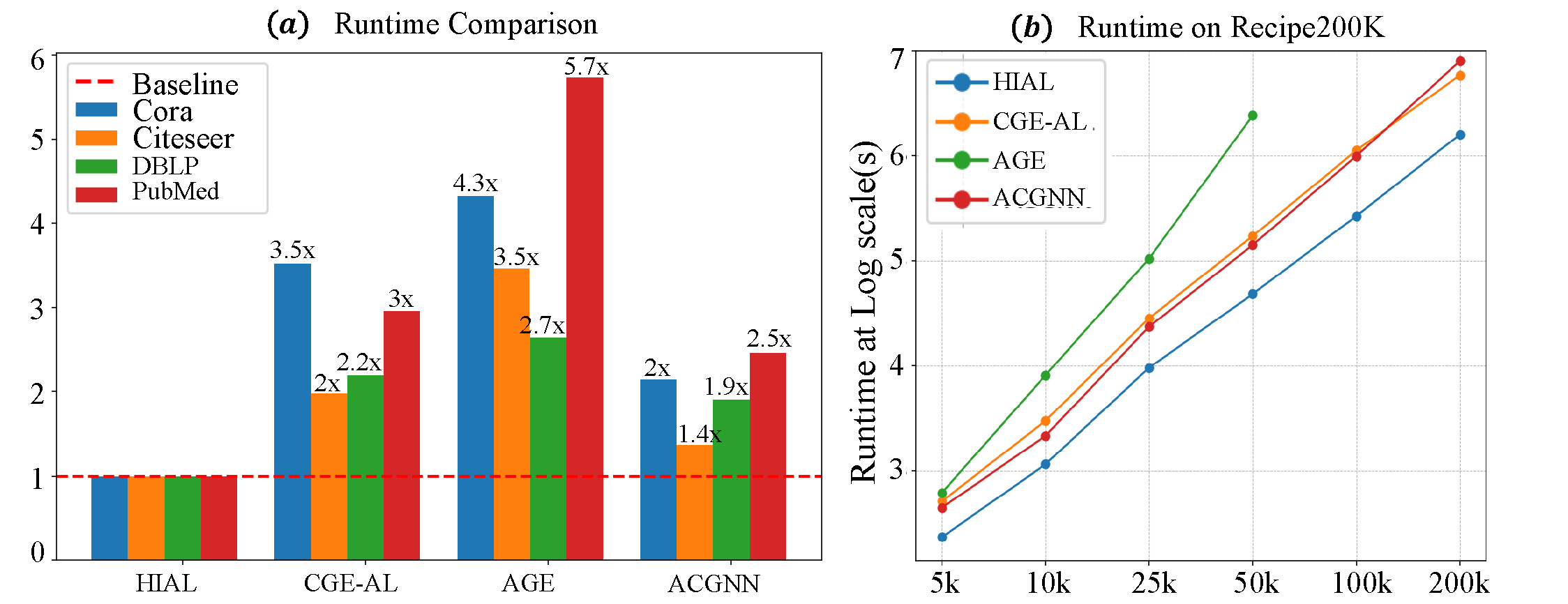}
    \caption{\((\mathbf{a})\) End-to-end runtime comparison on GPU and \((\mathbf{b})\) Runtime comparison on Recipe200K.}
    \label{efficiency}
\end{figure}
Another advantage of HIAL is its high efficiency in node selection. As an oracle-free and learning-free method, HIAL avoids the iterative delays required by learning-based approaches that must wait for oracle-provided labels. The end-to-end runtime comparison is presented in Figure \ref{efficiency}(a).\\
\textbf{\underline{Efficiency}} We compare HIAL with learning-based baselines, as heuristic and other learning-free baselines are structurally simple. As illustrated in Figure~\ref{efficiency}(a), HIAL consistently demonstrates superior efficiency across all datasets, establishing a baseline that other methods exceed.\\
\textbf{\underline{Scalability}} To test scalability, we sampled graphs of varying sizes from the large-scale Recipe200K dataset. Figure~\ref{efficiency}(b) shows that HIAL maintains the lowest runtime across all scales, proving its excellent scalability. In contrast, other methods like AGE record significantly longer runtimes, confirming HIAL's capability to handle large-scale graphs.

\subsection{Generalization(Q4)}
Beyond \(HOIP\) module, HIAL demonstrates strong adaptability to diverse HNNs. We validated this by replacing \(HOIP\) with the distinct message-passing mechanisms from representative HNNs like HGNN, UniGCN, VilLain, and AHGAE. We then evaluated these variants using a fixed budget of nodes selected by HIAL. The results are in Table \ref{Generalization}:\\
\textbf{\underline{Across HNN Architectures}} Regardless of the HNN’s propagation method, HIAL outperforms all baselines, which indicates the generality of HIAL across HNN architectures.\\
\textbf{\underline{\(HOIP\)’s Superiority}} \(HOIP\) captures hypergraph high - order interactions more effectively than conventional HNN propagation, enabling HIAL to achieve SOTA accuracy.

\begin{table}[t]
\centering
\caption{Test accuracy \% comparison on different HNN propagation methods.}
\begin{tabular}{c|ccccccc} 
\toprule
\multicolumn{1}{c|}{Method} & Cora & Citeseer & DBLP & PubMed\\
\midrule
HIAL-HGNN & 70.95 & \textcolor{orange}{60.20} & \textcolor{blue}{75.75} & \textcolor{blue}{75.69} \\
HIAL-UniGCN & \textcolor{blue}{72.71} & 60.13 & 75.19 & \textcolor{orange}{75.31}\\
HIAL-VilLain & 71.40 & \textcolor{blue}{60.26} & \textcolor{orange}{75.42} & 74.68\\
HIAL-AHGAE & \textcolor{orange}{72.03} & 60.05 & 74.85 & 74.93\\
\midrule
\textbf{HIAL} & \textcolor{red}{\textbf{73.72}} & \textcolor{red}{\textbf{62.76}} & \textcolor{red}{\textbf{77.30}} & \textcolor{red}{\textbf{76.66}}\\
\bottomrule
\end{tabular}
\label{Generalization}
\end{table}

\subsection{Robustness(Q5)}
The HIAL model demonstrates strong hyperparameter robustness. We tested parameters $\alpha$, $\beta$, and $\gamma$ across a range from \(0.1\) to \(0.9\), with detailed figures placed in the supplementary material due to space limitations.\\
\textbf{Parameter $\alpha$} On Cora dataset, accuracy is tightly bound within the \textbf{[\(0.71\), \(0.72\)]} range. On Citeseer, it is consistently maintained within \textbf{[\(0.58\), \(0.61\)]}. The maximum performance gap is only about 3\%.\\
\textbf{Parameter $\beta$} This parameter demonstrates the highest stability. On both datasets, accuracy curves are nearly flat, showing negligible variance across the parameter range.\\
\textbf{Parameter $\gamma$} The model again shows minimal sensitivity. For instance, accuracy varies by less than \(3\%\) on Cora, staying within the \textbf{[\(0.71\), \(0.73\)]} range.

\section{Conclusion}
We introduce HIAL, a novel influence maximization-based framework for hypergraph active learning, designed to address the limitation of existing methods that lose critical high-order information through destructive graph expansion. Our core contribution is to reformulate the HAL problem as an influence maximization task that operates natively on the hypergraph. To this end, we design a dual-perspective objective function that evaluates nodes in both the feature space and topological structure based on our novel high order-aware propagation mechanism. Then we employ an efficient greedy algorithm with a formal approximation guarantee. Extensive experiments on seven public datasets robustly demonstrate that HIAL outperforms state-of-the-art baselines in terms of performance, efficiency, generality, and robustness.

\bigskip
\bibliography{aaai2026}
\end{document}